\documentclass{article}
\usepackage{arxiv}

\usepackage[utf8]{inputenc}
\usepackage[T1]{fontenc}
\usepackage{graphicx}
\usepackage{booktabs}
\usepackage{amsmath,amssymb}
\usepackage{multirow}
\usepackage{subcaption}
\usepackage{hyperref}
\usepackage{tikz}
\usepackage{pgfplots}
\pgfplotsset{compat=1.18}
\usetikzlibrary{positioning,arrows.meta,calc,decorations.pathreplacing}

\newcommand{\Rsq}{R$^2$}

\newcommand{\etal}{\textit{et al.}}

\renewcommand{\headeright}{}
\renewcommand{\undertitle}{}

\begin{document}

\title{Do Foundation Models Know Geometry?\\
Probing Frozen Features for Continuous Physical Measurement}

\author{
  Yakov Pyotr Shkolnikov \\
  \texttt{yshkolni@gmail.com}
}

\maketitle

\begin{abstract}
Vision-language models encode continuous geometry that their text
pathway fails to express: a 6{,}000-parameter linear probe extracts
hand joint angles at 6.1$^\circ$ MAE from frozen features, while the
best text output achieves only 20.0$^\circ$---a 3.3$\times$
bottleneck. LoRA fine-tuning (r\,{=}\,16, 2{,}000 images) narrows
this gap to 6.5$^\circ$, providing evidence for a pathway-training
deficit rather than a representational one.
Training objective determines accuracy more than architecture:
five encoders spanning self-supervised, contrastive, and hybrid
paradigms converge to statistically equivalent accuracy
(R$^2$$\approx$0.55, TOST-equivalent at $\Delta$\,{=}\,0.03) despite
sharing as little as CKA\,{=}\,0.41 representational similarity---functional
convergence without representational convergence, extending the
platonic representation hypothesis to continuous geometric targets.
Results validated across fourteen backbones on head pose, rigid
objects, gaze, and camera intrinsics; rankings hold under nested
10-fold CV (Friedman $\chi^2$\,{=}\,94.3,
$p$\,${<}$\,10$^{-15}$).
\end{abstract}

\keywords{Foundation models \and Geometric probing \and Linear probes
\and Vision-language models \and Representation analysis}

% ============================================================
% 1. INTRODUCTION
% ============================================================
\section{Introduction}
\label{sec:intro}

Foundation models are increasingly deployed for quantitative visual tasks,
yet we lack systematic understanding of how well their representations encode
continuous physical measurements.
Practitioners prompt vision-language models for quantitative estimates and
receive imprecise answers with errors of 20--39$^\circ$
(Table~\ref{tab:main_results}).
Whether this reflects a fundamental limitation of the representation or
merely a bottleneck of the text interface remains an open question.

Fu~\etal~\cite{fu2025hidden} demonstrate that VLM visual representations
encode correct depth and correspondence information that the text generation
pathway fails to express.
Kodathala and Vunnam~\cite{kodathala2025bottleneck} find that 99.3\% of
visual samples suffer perceptual degradation when processed through text.
These studies diagnose the problem but do not offer a constructive solution
for continuous measurement.

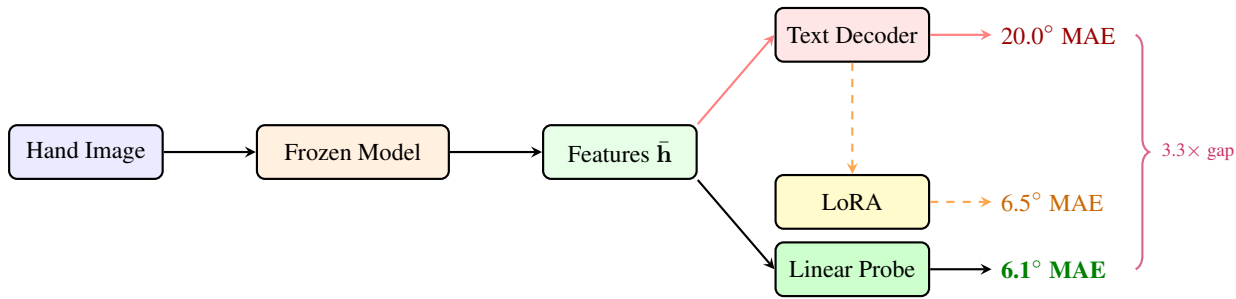
\begin{figure}[t]
\centering
\resizebox{\columnwidth}{!}{%
% Overview figure: Frozen Probing vs Text Pathway
\begin{tikzpicture}[
  >=stealth,
  node distance=1.0cm and 1.2cm,
  box/.style={draw, rounded corners=3pt, minimum height=0.7cm, minimum width=2.0cm,
              font=\small, align=center, thick},
  arrow/.style={->, thick},
  darrow/.style={->, thick, dashed},
]

% Input
\node[box, fill=blue!8] (img) {Hand Image};

% Foundation model
\node[box, fill=orange!12, right=1.2cm of img, minimum width=2.5cm] (model)
  {Frozen Model};

% Hidden features
\node[box, fill=green!10, right=1.2cm of model] (feat)
  {Features $\bar{\mathbf{h}}$};

% Text pathway (top -- higher error)
\node[box, fill=red!10, above right=0.8cm and 1.0cm of feat] (text)
  {Text Decoder};

% Probe pathway (bottom -- lower error)
\node[box, fill=green!20, below right=0.8cm and 1.0cm of feat] (probe)
  {Linear Probe};

% Outputs
\node[right=0.8cm of text, font=\small, red!60!black] (out_t)
  {20.0$^\circ$ MAE};

\node[right=0.8cm of probe, font=\small\bfseries, green!50!black] (out_p)
  {6.1$^\circ$ MAE};

% LoRA (near probe level -- visually represents recovery)
\node[box, fill=yellow!25, above=0.15cm of probe] (lora)
  {LoRA};

% Arrows
\draw[arrow] (img) -- (model);
\draw[arrow] (model) -- (feat);
\draw[arrow, red!50] (feat.north east) -- (text.west);
\draw[arrow] (feat.south east) -- (probe.west);
\draw[arrow, red!50] (text) -- (out_t);
\draw[arrow] (probe) -- (out_p);
\draw[darrow, orange!70] (text.south) -- (lora.north);
% LoRA output to the right
\node[right=0.8cm of lora, font=\small, orange!80!black] (out_l)
  {6.5$^\circ$ MAE};
\draw[darrow, orange!70] (lora) -- (out_l);

% Bottleneck brace (opening right toward label)
% Use a fixed x so different text widths don't tilt the brace
\path let \p1=(out_t.east), \p2=(out_p.east) in
  coordinate (brace top) at ({max(\x1,\x2)+4pt}, \y1)
  coordinate (brace bot) at ({max(\x1,\x2)+4pt}, \y2);
\draw[decorate, decoration={brace, amplitude=5pt}, thick, purple!60]
  (brace top) -- (brace bot)
  node[midway, right=6pt, font=\scriptsize, purple!80] {3.3$\times$ gap};

\end{tikzpicture}%
}
\caption{Overview. Frozen foundation model features encode continuous
geometry (joint angles) with 6.1$^\circ$ MAE via a linear probe, while
the text pathway achieves only 20.0$^\circ$, a 3.3$\times$
bottleneck. Adding LoRA fine-tuning (r\,{=}\,16) partially recovers
probe-level accuracy (6.5$^\circ$) through the text pathway.}
\label{fig:overview}
\end{figure}

We address this gap by systematically probing frozen features
of fourteen foundation models for continuous geometric
quantities (Fig.~\ref{fig:overview}).
Using four datasets spanning articulated and rigid pose
(FreiHAND hand images, BIWI head pose, YCB-Video object
pose, MPIIFaceGaze~\cite{zhang2017mpiifacegaze} gaze direction),
we test whether training methodology shapes geometric
encoding.
Our central finding is that \emph{training objective---not
architecture---determines geometric accuracy}, and that diverse
foundation models converge to equivalent geometric probing
despite representationally dissimilar features.

Our three contributions:
\begin{enumerate}
  \item \textbf{The text bottleneck is a pathway-training deficit,
  not a representation deficit.}
  Frozen probes achieve 6.1$^\circ$ MAE while text output achieves
  only 20.0$^\circ$, a 3.3$\times$ gap.
  LoRA (r\,{=}\,16, 2{,}000 images) narrows this to 6.5$^\circ$,
  providing evidence that geometry is encoded but not routed through
  the text pathway.
  Layer-wise probing shows LoRA preserves geometric signal at
  layers where the frozen base loses it (Appendix~\ref{sec:supp_lora_layers}).
  \item \textbf{Training objective determines accuracy
  more than architecture.}
  A controlled ablation (DeiT3-L vs.\ ConvNeXt-L, matched
  IN-1K pretraining) shows no ViT advantage
  (R$^2$\,{=}\,0.38 vs.\ 0.41); the 0.15 gap to the
  cluster reflects self-supervised/contrastive pretraining,
  not attention mechanisms.
  Five encoders converge to \Rsq{}$\,{\approx}\,$0.55 via
  representationally dissimilar features (CKA as low as 0.41),
  demonstrating functional convergence without representational
  convergence.
  \item \textbf{Geometry is spatially task-dependent.}
  Patch ablation drops head-pose \Rsq{} by 0.13 (loosely-framed faces)
  but object-pose by only 0.003 (tightly-cropped), explaining
  cross-dataset variation in attention pooling gains.
\end{enumerate}
These findings enable a single frozen backbone to function as a multi-task
geometric probe. Hand pose, head pose, object pose, and camera
intrinsics are all linearly readable, with each task adding
${\sim}$6{,}000 probe parameters.

% ============================================================
% 2. RELATED WORK
% ============================================================
\section{Related Work}
\label{sec:related}

\paragraph{Text bottleneck in VLMs.}
Fu~\etal~\cite{fu2025hidden} show that VLM visual features encode depth and
correspondence that text generation discards (21.7\% and 45.5\% degradation
respectively).
Kodathala and Vunnam~\cite{kodathala2025bottleneck} systematically document
this gap across perceptual tasks.
Guo~\etal~\cite{guo2025flatlands} independently identify the same
discrete-tokenizer bottleneck and propose architectural modifications
with direct regression heads.
The GIQ benchmark~\cite{giq2025} independently shows VLMs achieve
below 20\% on geometric shape reasoning via text, further evidence
that text pathways discard geometric detail.
G2VLM~\cite{g2vlm2025} proposes geometry-grounded VLMs for spatial tasks.
We complement these by demonstrating that frozen features already encode
the geometry, and LoRA fine-tuning recovers it through layer-wise
preservation (see Appendix~\ref{sec:supp_lora_layers}) without architectural changes.

\paragraph{Probing foundation models for 3D awareness.}
El~Banani~\etal~\cite{elbanani2024probe3d} evaluate 3D awareness of visual
encoders on depth, surface normals, and correspondence.
Liu~\etal~\cite{lexicon3d} map the representational space of vision
models for 3D understanding across multiple probe types.
Zhan~\etal~\cite{zhan2024probing} probe video models for physical properties.
Yao~\etal~\cite{yao2025reading} show that latent probes detect OCR errors
invisible to text output.
Yue~\etal~\cite{yue2024fit3d} improve 2D features via 3D-aware
fine-tuning, focusing on dense prediction.
Kar~\etal~\cite{kar2024brave} systematically broaden the visual encoding
of VLMs and reveal significant variation in how different encoders
preserve visual information.
Tong~\etal~\cite{tong2024eyes} expose systematic visual shortcomings
of multimodal LLMs on basic perceptual patterns.
Chen~\etal~\cite{chen2025feat2gs} probe frozen features via Gaussian
splatting, disentangling geometry from texture.
These works focus on dense (per-pixel) or categorical probing.
We complement them with global (image-level) probing of
continuous scalar quantities.

\paragraph{Probing neural representations.}
Linear probing originates in NLP, where
Alain and Bengio~\cite{alain2017probing} introduced linear classifier
probes to interpret intermediate layers, and
Conneau~\etal~\cite{conneau2018probing} systematically probed sentence
embeddings for linguistic properties.
Hewitt and Liang~\cite{hewitt2019control} show that probe complexity must
be controlled to distinguish learned representations from probe
memorization---our use of reduced-rank regression (rank 3--8)
addresses this concern.
Basile~\etal~\cite{basile2025headpursuit} concurrently probe attention-head
specialization in multimodal transformers, showing that editing
${\sim}1\%$ of heads can reliably steer model outputs.
We extend this NLP probing tradition to continuous geometric targets
across vision and vision-language models.

\paragraph{Geometric regression from VLMs.}
SpatialVLM~\cite{spatialvlm} trains VLMs for spatial reasoning through
chain-of-thought.
Xue~\etal~\cite{xue2024reovlm} fine-tune VLMs for rigid object pose.
For hand pose specifically, HaMeR~\cite{pavlakos2024hamer} and
Hamba~\cite{dong2024hamba} achieve 5.7 and 5.3\,mm PA-MPVPE
respectively using MANO-based mesh recovery---a fundamentally
different approach measuring positional error (mm) rather than
angular error (degrees).
Our approach differs in using frozen features with lightweight probes rather
than end-to-end fine-tuning, enabling direct comparison across architectures.

% ============================================================
% 3. METHOD
% ============================================================
\section{Method}
\label{sec:method}

\subsection{Problem Setup}
\label{sec:setup}

Given an image $\mathbf{x}_i$ and a frozen model $f$, we extract hidden
activations $\mathbf{H}_i^{(\ell)} \in \mathbb{R}^{T \times d}$ at layer
$\ell$ (where $T$ is the sequence length and $d$ the hidden dimension).
We mean-pool spatially to obtain a global feature vector
$\bar{\mathbf{h}}_i = \frac{1}{T'}\sum_{t \in \mathcal{P}}
\mathbf{H}_{i,t}^{(\ell)}$, where $\mathcal{P}$ excludes special tokens
(CLS, registers) for models that use them; CLIP and all
VLM encoders pool all tokens including CLS.
A linear probe $\hat{\mathbf{y}}_i = \mathbf{W}\bar{\mathbf{h}}_i +
\mathbf{b}$ maps features to continuous targets
$\mathbf{y}_i \in \mathbb{R}^K$ (joint angles in degrees).

We use reduced-rank ridge regression
(RRR;~\cite{izenman1975rrr}): fit Ridge($\alpha$) then truncate the
weight matrix via SVD to rank $r$.
Hyperparameters are swept over
$r \in \{3,4,5,6,8\}$ and $\alpha \in \{1,10,100,1000\}$.
For each model, we select the layer maximizing hold-out \Rsq{};
nested 10-fold CV confirms that this selection does not change
rankings (cluster models within 0.006; see Limitations M1).
We report nested CV as the primary metric and
hold-out results in the appendix.

\subsection{Datasets}
\label{sec:datasets}

\textbf{FreiHAND}~\cite{freihand}: 32{,}560 hand images with 21 3D
keypoints from 32 subjects.
We compute per-finger mean flexion angles (3 joints per finger, 5 fingers)
and use a fixed 8{,}000-image subset (6{,}400 train / 1{,}600 test;
indices 0--32{,}559 only, excluding augmented copies).

\textbf{BIWI}~\cite{biwi}: 15{,}678 RGBD head images from 20 subjects with
yaw, pitch, and roll labels.
We use subject-stratified splits (16 train / 4 test subjects).

\textbf{YCB-Video}~\cite{xiang2018ycbvideo}: 133{,}827 frames of 21
objects with 6DoF pose.
We subsample 900 images and probe rotation (Euler angles) and translation
separately.

\subsection{Models}
\label{sec:models}

We evaluate fourteen models spanning four training approaches:

\begin{itemize}
  \item \textbf{Self-supervised}: DINOv2 ViT-L~\cite{dinov2}
  (ViT-L/14~\cite{dosovitskiy2021vit}),
  DINOv3 ViT-L~\cite{dinov3}, DINOv2 ViT-B
  \item \textbf{Contrastive VL}: CLIP ViT-L~\cite{clip},
  SigLIP ViT-L~\cite{siglip}, SigLIP-B
  \item \textbf{Hybrid VL}: SigLIP~2 ViT-L~\cite{siglip2},
  InternViT-300M~\cite{chen2024internvl,chen2024internvl15}
  \item \textbf{Generative VLMs}: Qwen2.5-VL-3B, Qwen2.5-VL-7B~\cite{wang2024qwen2vl},
  QwenVIT-3B, QwenVIT-merger, Gemma~3 4B-IT~\cite{gemma3}
  \item \textbf{CNN baseline}: ConvNeXt-L~\cite{liu2022convnext} (IN-22K+1K)
\end{itemize}

For VLMs, we extract from LLM decoder layers using a fixed prompt
(``Describe this hand.'').
For vision-only models, we extract from intermediate transformer blocks.
Code for all extractors, probes, and statistical tests will be released
open-source.

\subsection{Evaluation}
\label{sec:eval}

We report MAE (degrees) and \Rsq{} (uniform mean across 5 finger targets).
Statistical comparisons use TOST equivalence testing
($\Delta$\,{=}\,0.03, chosen as a practically meaningful threshold:
models differing by ${<}$0.03 \Rsq{} are interchangeable for
deployment)~\cite{schuirmann1987tost,lakens2017equivalence},
Friedman rank tests~\cite{friedman1937,demsar2006statistical},
and BCa bootstrap confidence intervals (10{,}000 resamples,
bias-corrected and accelerated~\cite{efron1987bca}).

% ============================================================
% 4. RESULTS
% ============================================================
\section{Results}
\label{sec:results}

\subsection{Main Results: Probe vs.\ Text}
\label{sec:main_results}

Table~\ref{tab:main_results} compares frozen probes against text and
task-specific baselines.
The best frozen probe (SigLIP~2, L16) achieves 6.14$^\circ$ MAE on FreiHAND
hand joint angles, while the best text baseline (few-shot prompting of
Qwen-3B) achieves 20.0$^\circ$, a 2.7$\times$ within-model gap
(3.3$\times$ vs.\ the best probe).
Even MediaPipe Hands~\cite{zhang2020mediapipe}, a dedicated hand pose
model (3.7M parameters), achieves only 16.3$^\circ$ when evaluated
zero-shot via its 3D world landmarks (caveat: monocular depth
estimation is less accurate than FreiHAND's multi-view ground truth;
see Discussion).
Chain-of-thought prompting worsens performance (139.3$^\circ$),
as the multi-step reasoning produces hallucinated angular values often
exceeding the anatomical range.

\begin{table}[t]
\centering
\caption{Four readout regimes for geometric information on FreiHAND.
MAE in degrees; \Rsq{} is uniform mean across 5 finger targets.
LoRA: r\,{=}\,16, 2{,}000 images, 2 epochs.
Frozen probes: RRR, 6{,}400 images.
$^*$MediaPipe uses monocular 3D world landmarks evaluated zero-shot
(see Discussion for caveats).}
\label{tab:main_results}
\begin{tabular}{llccc}
\toprule
Regime & Method & MAE ($^\circ$) & \Rsq{} & Parse \\
\midrule
Task-specific
  & MediaPipe Hands$^*$~\cite{zhang2020mediapipe} & 16.3 & $-$2.44 & N/A \\
\cmidrule{1-5}
\multirow{3}{*}{Text generation}
  & Direct prompt (Qwen-3B) & 39.3 & --- & varies \\
  & Chain-of-thought (Qwen-3B) & 139.3 & --- & varies \\
  & Few-shot 3-ex.\ (Qwen-3B) & 20.0 & --- & varies \\
\cmidrule{1-5}
\multirow{2}{*}{LoRA text}
  & LoRA Qwen-3B & 7.45 & 0.299 & 100\% \\
  & LoRA Gemma~3 4B & \textbf{6.51} & 0.400 & 100\% \\
\cmidrule{1-5}
\multirow{3}{*}{Frozen probe}
  & RRR (Qwen-3B L11) & 7.28 & 0.435 & N/A \\
  & RRR (Gemma~3 L0) & 6.59 & 0.505 & N/A \\
  & RRR (SigLIP~2 L16) & \textbf{6.14} & \textbf{0.559} & N/A \\
\bottomrule
\end{tabular}
\end{table}

\subsection{LoRA Fine-Tuning Narrows the Text Bottleneck}
\label{sec:lora}

We test whether lightweight fine-tuning can teach the text pathway to
read the geometry encoded in frozen features.
Table~\ref{tab:main_results} presents results for LoRA~\cite{hu2022lora} fine-tuning (r\,{=}\,16,
2{,}000 training images, 2 epochs) on two model families.

Gemma~3 LoRA achieves 6.51$^\circ$ MAE, surpassing its own frozen probe
(6.59$^\circ$) with 3.2$\times$ less training data.
Qwen-3B LoRA achieves 7.45$^\circ$, matching its frozen probe (7.28$^\circ$).
Both achieve 100\% parse rates.

\Rsq{} recovery is partial and model-dependent: 79\% for Gemma~3
(0.400/0.505) and 69\% for Qwen-3B (0.299/0.435).
The MAE advantage with lower \Rsq{} reflects the error distribution of
text-generation predictions: quantized to 0.1$^\circ$ resolution with
occasional large outliers (6.4\% for Gemma~3, 9.1\% for Qwen-3B
exceed 20$^\circ$ error).
MAE is insensitive to such outliers while \Rsq{} (MSE-based) is not.

These results provide evidence that LoRA enables the autoregressive
decoder to route existing geometric signals through the text pathway.
The frozen backbone is the sensor and LoRA is the readout interface.

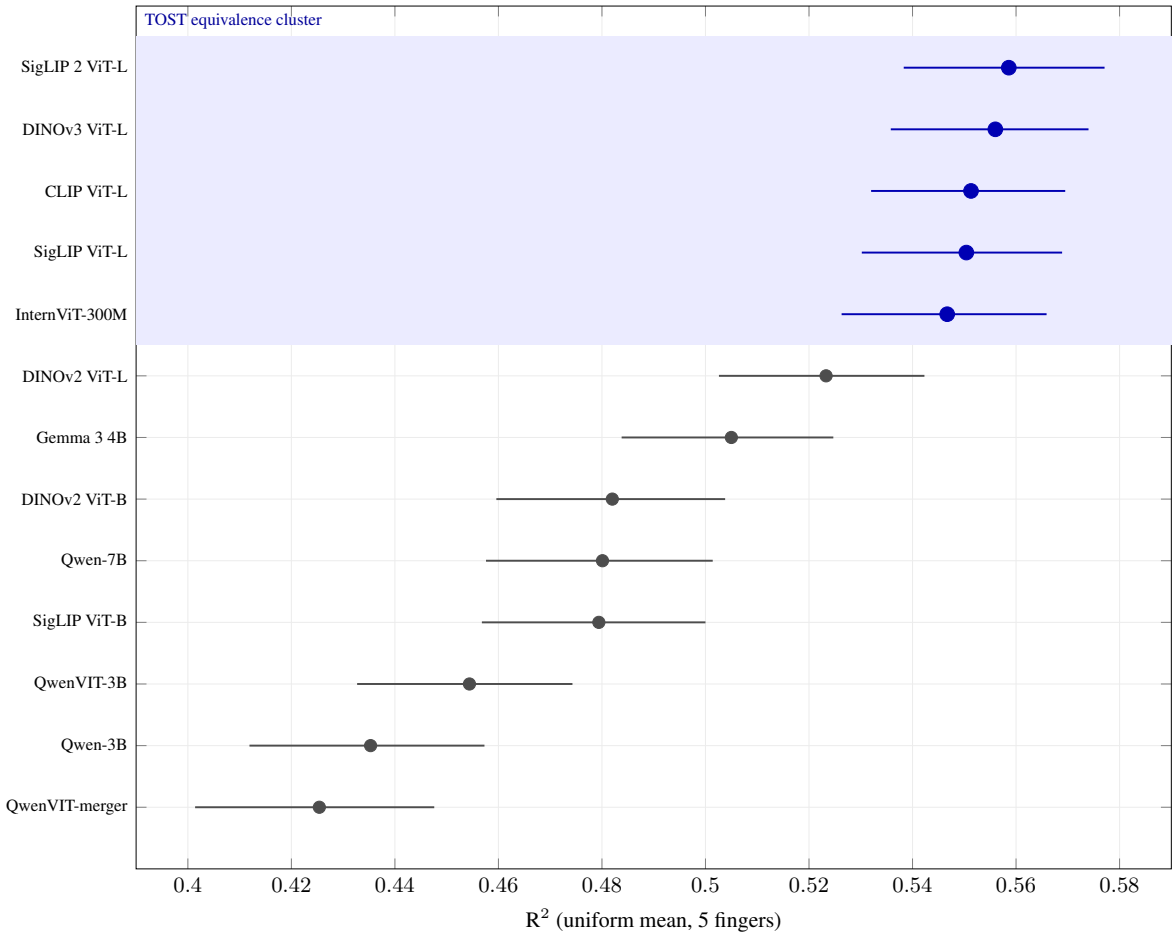
\begin{figure}[t]
\centering
\resizebox{0.95\columnwidth}{!}{%
\begin{tikzpicture}
\begin{axis}[
  width=\columnwidth,
  height=0.85\columnwidth,
  xlabel={\Rsq{} (uniform mean, 5 fingers)},
  xmin=0.39, xmax=0.59,
  ymin=0, ymax=14,
  ytick={1,2,3,4,5,6,7,8,9,10,11,12,13},
  yticklabels={QwenVIT-merger,Qwen-3B,QwenVIT-3B,SigLIP ViT-B,Qwen-7B,DINOv2 ViT-B,Gemma 3 4B,DINOv2 ViT-L,InternViT-300M,SigLIP ViT-L,CLIP ViT-L,DINOv3 ViT-L,SigLIP 2 ViT-L},
  y tick label style={font=\scriptsize, anchor=east},
  x tick label style={font=\small},
  label style={font=\small},
  grid=major,
  grid style={gray!15},
  clip=false,
]

% Equivalence cluster shading
\fill[blue!8] (axis cs:0.39,8.5) rectangle (axis cs:0.59,13.5);

% CI lines and point estimates
\addplot[black!70, thick] coordinates {(0.4014,1) (0.4476,1)};
\addplot[only marks, mark=*, mark size=2.5pt, black!70] coordinates {(0.4254,1)};
\addplot[black!70, thick] coordinates {(0.4119,2) (0.4573,2)};
\addplot[only marks, mark=*, mark size=2.5pt, black!70] coordinates {(0.4353,2)};
\addplot[black!70, thick] coordinates {(0.4327,3) (0.4743,3)};
\addplot[only marks, mark=*, mark size=2.5pt, black!70] coordinates {(0.4544,3)};
\addplot[black!70, thick] coordinates {(0.4568,4) (0.5000,4)};
\addplot[only marks, mark=*, mark size=2.5pt, black!70] coordinates {(0.4794,4)};
\addplot[black!70, thick] coordinates {(0.4576,5) (0.5014,5)};
\addplot[only marks, mark=*, mark size=2.5pt, black!70] coordinates {(0.4801,5)};
\addplot[black!70, thick] coordinates {(0.4596,6) (0.5038,6)};
\addplot[only marks, mark=*, mark size=2.5pt, black!70] coordinates {(0.4820,6)};
\addplot[black!70, thick] coordinates {(0.4838,7) (0.5247,7)};
\addplot[only marks, mark=*, mark size=2.5pt, black!70] coordinates {(0.5050,7)};
\addplot[black!70, thick] coordinates {(0.5026,8) (0.5423,8)};
\addplot[only marks, mark=*, mark size=2.5pt, black!70] coordinates {(0.5233,8)};
\addplot[blue!70!black, thick] coordinates {(0.5263,9) (0.5659,9)};
\addplot[only marks, mark=*, mark size=3pt, blue!70!black] coordinates {(0.5467,9)};
\addplot[blue!70!black, thick] coordinates {(0.5302,10) (0.5689,10)};
\addplot[only marks, mark=*, mark size=3pt, blue!70!black] coordinates {(0.5504,10)};
\addplot[blue!70!black, thick] coordinates {(0.5320,11) (0.5695,11)};
\addplot[only marks, mark=*, mark size=3pt, blue!70!black] coordinates {(0.5513,11)};
\addplot[blue!70!black, thick] coordinates {(0.5358,12) (0.5740,12)};
\addplot[only marks, mark=*, mark size=3pt, blue!70!black] coordinates {(0.5560,12)};
\addplot[blue!70!black, thick] coordinates {(0.5383,13) (0.5771,13)};
\addplot[only marks, mark=*, mark size=3pt, blue!70!black] coordinates {(0.5586,13)};

% Cluster label
\node[font=\scriptsize, blue!60!black, anchor=south west] at (axis cs:0.39,13.5) {TOST equivalence cluster};

\end{axis}
\end{tikzpicture}%
}
\caption{Bootstrap 95\% CIs for 13 models on FreiHAND
(ConvNeXt-L omitted; see Table~\ref{tab:cross_arch}).
Five models form a TOST equivalence cluster (shaded) at
\Rsq{}$\,{\approx}\,$0.55. DINOv2 falls outside despite being
the same architecture family as DINOv3.}
\label{fig:forest}
\end{figure}

\subsection{Cross-Architecture Comparison}
\label{sec:cross_arch}

Table~\ref{tab:cross_arch} and Fig.~\ref{fig:forest} show \Rsq{} for
all fourteen models on FreiHAND.
Five vision encoders form a statistical equivalence cluster at
\Rsq{}$\,{\approx}\,$0.55 (TOST-equivalent, all 10 pairwise tests pass,
$\Delta$\,{=}\,0.03): SigLIP~2 (0.559), DINOv3 (0.556), CLIP (0.551),
SigLIP (0.550), and InternViT (0.547).
DINOv2 (0.523) falls outside this cluster despite being the same
architecture family as DINOv3.

Autoregressive LLM processing reduces hand-pose accuracy
(Gemma~3 L0: 0.505, Qwen-3B: 0.435). Text-generation
preparation is associated with reduced encoding of articulated geometry.
This degradation is task-dependent and dissolves on rigid objects
(all $\approx$0.70 on YCB-Video). The LLM pathway
preserves coarse pose but discards fine-grained joint angles.
ViT-B base models nearly match ViT-L (DINOv2-B: 0.482, SigLIP-B: 0.479),
suggesting geometric encoding is not primarily capacity-limited.
ConvNeXt-L (0.455) falls 0.10 below the ViT cluster, but a controlled
ablation (Sec.~\ref{sec:controlled}) shows this reflects pretraining,
not architecture.

\begin{table}[t]
\centering
\caption{Cross-architecture comparison on FreiHAND (8{,}000 images).
\Rsq{} is uniform mean across 5 finger targets.
$^\dagger$Nested 10-fold CV (Friedman $\chi^2$\,{=}\,94.3,
$p$\,${<}$\,10$^{-15}$; Nemenyi CD\,{=}\,4.45).
Cluster models within 0.012 of test \Rsq{}.
--- = not included in CV (added for ablation/scaling analysis).}
\label{tab:cross_arch}
\begin{tabular}{llccc}
\toprule
Model & Training & Layer & \Rsq{} & CV \Rsq{}$^\dagger$ \\
\midrule
SigLIP~2 ViT-L & Hybrid VL & L16 & \textbf{0.559} & 0.563 \\
DINOv3 ViT-L & Self-supervised & L20 & 0.556 & 0.550 \\
CLIP ViT-L & Contrastive VL & L20 & 0.551 & 0.554 \\
SigLIP ViT-L & Contrastive VL & L16 & 0.550 & 0.549 \\
InternViT-300M & Hybrid VL & L20 & 0.547 & 0.549 \\
\midrule
DINOv2 ViT-L & Self-supervised & L20 & 0.523 & 0.494 \\
Gemma~3 4B-IT & Generative VLM & L0 & 0.505 & 0.512 \\
DINOv2 ViT-B & Self-supervised & L12 & 0.482 & --- \\
Qwen2.5-VL-7B & Generative VLM & L8 & 0.480 & 0.480 \\
SigLIP ViT-B & Contrastive VL & L12 & 0.479 & --- \\
ConvNeXt-L & Supervised CNN & S2 & 0.455 & --- \\
QwenVIT-3B & Vision enc.\ only & L24 & 0.454 & 0.460 \\
Qwen2.5-VL-3B & Generative VLM & L11 & 0.435 & 0.434 \\
QwenVIT-merger & Vision enc.\ only & --- & 0.425 & 0.431 \\
\bottomrule
\end{tabular}
\end{table}

\subsection{Cross-Dataset Validation}
\label{sec:cross_dataset}

\paragraph{BIWI head pose.}
On BIWI, the FreiHAND equivalence cluster dissolves: DINOv3 leads
(\Rsq{}\,{=}\,0.607), followed by DINOv2 (0.532) and SigLIP~2 (0.455),
confirming that rankings are task-dependent.
Pitch is best predicted (\Rsq{}\,{=}\,0.948) and roll hardest
(\Rsq{}\,{=}\,0.168).
Attention pooling produces large gains: DINOv2 jumps from 0.532 to
0.892, with roll rising from 0.052 to 0.779.
This 0.36 \Rsq{} gain reflects spatial concentration of head-pose
information in face patches within loosely-framed images.

\paragraph{YCB-Video object pose.}
All models achieve $\approx$0.70 rotation \Rsq{} on YCB-Video, with no
significant pairwise differences.
The autoregressive degradation observed on hands dissolves on rigid objects.
AttentionPool provides no benefit ($\Delta$\,{=}\,$-$0.06 to 0.00),
consistent with geometry being distributed across all patches in
tightly-cropped object images.

\paragraph{Gaze direction.}
On MPIIFaceGaze~\cite{zhang2017mpiifacegaze} (45{,}000 face images), DINOv3 dominates
(\Rsq{}\,{=}\,0.787, 3.14$^\circ$ MAE), 0.21 above DINOv2.
Rankings differ from hand pose, and the best encoder varies
by task geometry (see Appendix~\ref{sec:supp_gaze} for full results).

\paragraph{Per-bone and camera intrinsics.}
Per-joint probing reveals a universal proximal-distal gradient
(MCP: 0.544, PIP: 0.559, DIP: 0.271), independent of model.
Frozen features also encode camera intrinsics
(\Rsq{}\,{=}\,0.81--0.94 for focal length), extending the
multi-task geometric probe beyond pose (full results in Appendix~\ref{sec:supp_intrinsics}).

\subsection{Controlled Architecture Ablation}
\label{sec:controlled}

To disentangle architecture from pretraining, we compare DeiT3-L~\cite{touvron2022deit3}
(ViT-L/16, 304M, ImageNet-1K supervised) against ConvNeXt-L~\cite{liu2022convnext}
(CNN, 198M, ImageNet-1K supervised).
With matched pretraining, the CNN slightly outperforms the ViT
(\Rsq{}\,{=}\,0.405 vs.\ 0.379).
Scaling ConvNeXt pretraining data from IN-1K to IN-22K lifts
\Rsq{} from 0.405 to 0.455 (+0.050), exceeding the architecture
effect ($-$0.026).
Both supervised-only models fall 0.15 below the self-supervised/contrastive
cluster. Geometric encoding quality is driven primarily by
training objective, consistent with controlled comparisons showing
that data and training signal dominate architecture~\cite{liu2025clipvsdino}.

% ============================================================
% 5. WHERE AND HOW GEOMETRY LIVES
% ============================================================
\section{Where and How Geometry Lives}
\label{sec:mechanistic}

The preceding sections establish \emph{what} frozen features encode.
We now investigate \emph{how} geometric information is organized within
these representations.

\subsection{Functional Convergence Without Representational Similarity}
\label{sec:cka}

\begin{figure}[t]
\centering
\resizebox{0.95\columnwidth}{!}{%
\begin{tikzpicture}
\begin{axis}[
  width=\columnwidth,
  height=0.65\columnwidth,
  xlabel={Linear CKA similarity},
  ylabel={$|\Delta$\Rsq{}$|$ between model pair},
  xmin=0.35, xmax=0.95,
  ymin=-0.003, ymax=0.085,
  grid=major,
  grid style={gray!20},
  tick label style={font=\small},
  label style={font=\small},
  legend style={font=\footnotesize, at={(0.02,0.98)}, anchor=north west},
]

% 15 ViT-L pairwise points (filled circles)
\addplot[only marks, mark=*, mark size=2.5pt, color=black!70] coordinates {
  (0.8810, 0.0330)
  (0.7030, 0.0270)
  (0.5230, 0.0360)
  (0.5650, 0.0280)
  (0.7080, 0.0240)
  (0.6910, 0.0060)
  (0.5220, 0.0030)
  (0.5580, 0.0050)
  (0.6810, 0.0090)
  (0.5260, 0.0090)
  (0.5240, 0.0010)
  (0.7440, 0.0030)
  (0.4120, 0.0080)
  (0.6310, 0.0120)
  (0.5540, 0.0040)
};
\addlegendentry{ViT-L pairs ($n$\,{=}\,15)}

% 13 cross-scale pairs involving ViT-B models (open triangles)
\addplot[only marks, mark=triangle*, mark size=3pt, color=teal!80!black, fill=teal!30] coordinates {
  (0.8610, 0.0410)
  (0.8000, 0.0740)
  (0.6590, 0.0680)
  (0.4270, 0.0770)
  (0.4700, 0.0690)
  (0.5940, 0.0650)
  (0.6590, 0.0440)
  (0.6430, 0.0770)
  (0.7710, 0.0710)
  (0.4930, 0.0800)
  (0.5100, 0.0720)
  (0.7360, 0.0680)
  (0.5890, 0.0030)
};
\addlegendentry{ViT-B pairs ($n$\,{=}\,13)}

% Highlight: highest CKA pair (DINOv2-DINOv3)
\addplot[only marks, mark=*, mark size=4pt, color=red!80!black, fill=red!30] coordinates {
  (0.881, 0.033)
};
\node[font=\scriptsize, anchor=south west, color=red!80!black] at (axis cs:0.881,0.034) {DINOv2--DINOv3};

% Highlight: lowest CKA pair (SigLIP2-CLIP)
\addplot[only marks, mark=*, mark size=4pt, color=blue!80!black, fill=blue!30] coordinates {
  (0.412, 0.008)
};
\node[font=\scriptsize, anchor=north east, color=blue!80!black] at (axis cs:0.410,0.007) {SigLIP\,2--CLIP};

% Correlation annotation
\node[font=\footnotesize, anchor=south east, text=black!60] at (axis cs:0.93,0.001) {$\rho$\,{=}\,0.03, $p$\,{=}\,0.88 ($n$\,{=}\,28)};

\end{axis}
\end{tikzpicture}%
}
\caption{CKA similarity vs.\ probing accuracy difference for all 28
pairwise comparisons among eight models (six ViT-L + two ViT-B) on FreiHAND.
Spearman $\rho$\,{=}\,0.03 ($p$\,{=}\,0.88):
no detectable correlation between representational
similarity and geometric probing accuracy ($n$\,{=}\,28).
The most similar pair (DINOv2--DINOv3, CKA\,{=}\,0.88) differs by 0.033
\Rsq{}; the least similar pair (SigLIP\,2--CLIP, CKA\,{=}\,0.41)
differs by only 0.008.}
\label{fig:cka_scatter}
\end{figure}
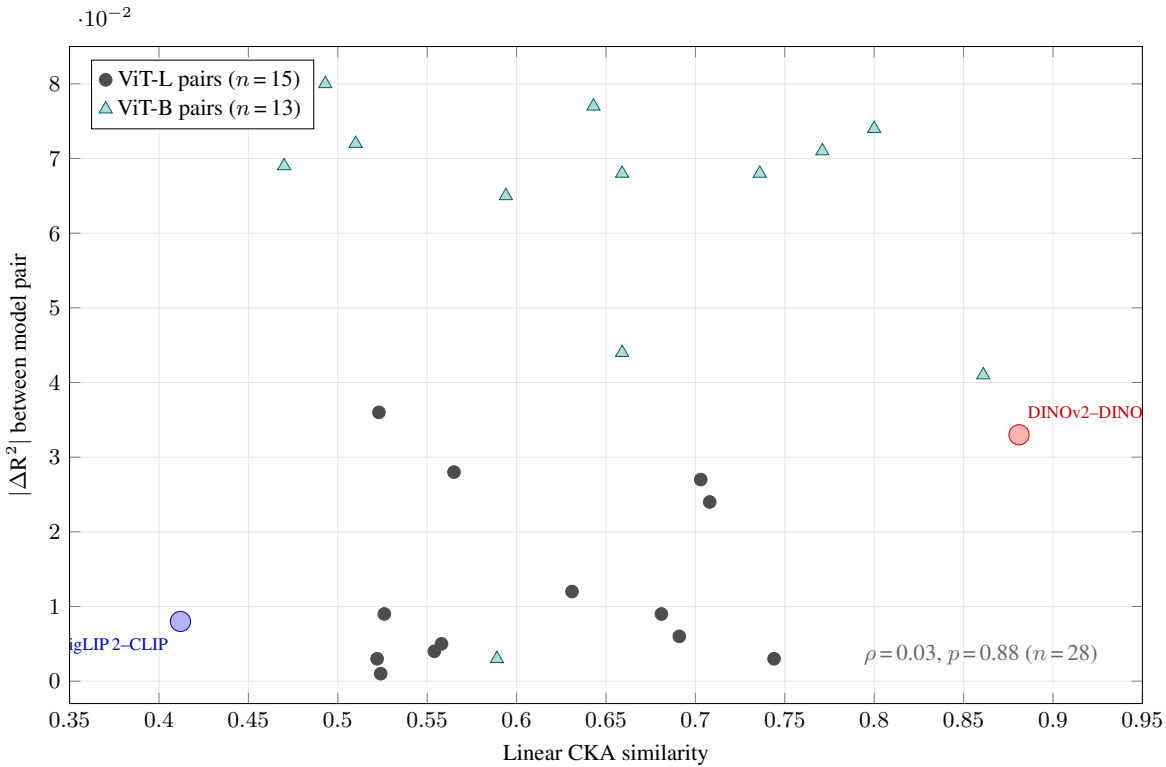

Critically, linear CKA~\cite{kornblith2019similarity}
(noting recent reliability concerns~\cite{davari2023cka_reliability})
analysis reveals that the
\Rsq{}$\,{\approx}\,$0.55 equivalence cluster reflects \emph{functional}
convergence rather than representational alignment (Fig.~\ref{fig:cka_scatter}).
Across all 28 pairwise comparisons among eight models (six ViT-L plus
DINOv2-B and SigLIP-B),
CKA similarity shows no detectable correlation with probing accuracy difference
(Spearman $\rho$\,{=}\,0.03, $p$\,{=}\,0.88).
DINOv2 and DINOv3 share CKA\,{=}\,0.881 yet differ by 0.033 \Rsq{},
while SigLIP~2 and CLIP share only CKA\,{=}\,0.412 yet differ by 0.008.
Multiple representational strategies converge on a shared geometric
readout, extending the platonic representation
hypothesis~\cite{huh2024platonic}: functional convergence exists but
does not require representational convergence, suggesting a \emph{weak}
form of the hypothesis.
This contrasts with the stronger representational alignment observed
across scientific foundation
models~\cite{edamadaka2025converging,platonic_universe2025}.
A simple dimensionality argument may partially explain why: the target
space is 5-dimensional while feature spaces are 1024-dimensional, so
many distinct linear projections can achieve similar regression accuracy.
Functional convergence may partly reflect the geometric inevitability of
projecting high-dimensional features onto low-dimensional targets.

\subsection{Layer Trajectory and Proximal-Distal Gradient}
\label{sec:layers}

\begin{figure}[t]
\centering
  \includegraphics[width=\columnwidth]{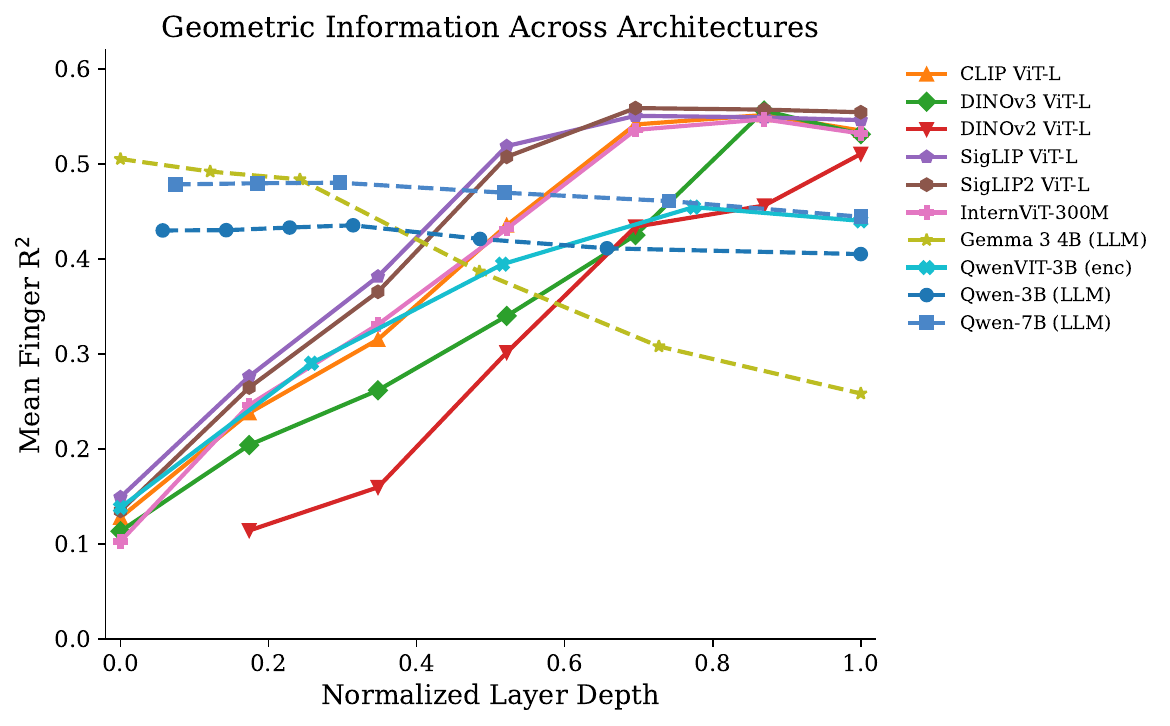}
  \caption{Layer-wise \Rsq{} on FreiHAND for ten models
  (solid: vision encoders; dashed: LLM decoders).
  X-axis is normalized layer depth (0\,{=}\,first, 1\,{=}\,last).
  Vision encoders rise monotonically; LLM decoders peak at early
  layers and decline, consistent with autoregressive processing
  discarding fine-grained geometry.}
  \label{fig:layer_curves}
\end{figure}

Layer-wise probing (Fig.~\ref{fig:layer_curves}) reveals
a proximal-distal gradient (PIP${>}$MCP${>}$DIP) at
every layer depth across all models.
Geometric signal builds monotonically from
R$^2$$\approx$0.11--0.28 at L4 to a peak of
$\approx$0.55 at L16--L20, declining slightly at the final layer.
Self-supervised models show delayed geometric emergence
compared to contrastive models; self-distillation may
concentrate geometric representations in deeper layers.
In contrast, VLM decoder layers (dashed in Fig.~\ref{fig:layer_curves})
peak at early layers and decline monotonically, consistent with
autoregressive processing discarding articulated geometry
(Gemma~3 peaks at L0, Qwen-3B at L11).
QwenVIT (vision encoder only) rises like other ViTs,
confirming that the decline is specific to LLM processing.

\subsection{Spatial Concentration and Patch Ablation}
\label{sec:patches}

Geometry is an ensemble property of attention: all 16 heads in DINOv2-L
carry comparable geometric signal (\Rsq{}\,{=}\,0.40--0.48 per head;
no joint specialization---maximum absolute Spearman correlation
between any head's attention entropy and any joint angle is
$|\rho|$\,{=}\,0.28; see Appendix~\ref{sec:supp_attn_heads}).
Patch ablation provides ablation evidence for task-dependent spatial
concentration.
Removing the 100 highest-norm patches from DINOv3 features drops BIWI
\Rsq{} by 0.126, while the same ablation on YCB-Video changes \Rsq{} by
only $-$0.003.
Random patch removal of equal count produces smaller BIWI drops ($-$0.107)
and larger YCB drops ($-$0.040). Geometric information is
specifically concentrated in high-activation patches for loosely-framed
subjects but distributed for tightly-cropped objects.
This explains why AttentionPool improves BIWI by +0.23--0.36 \Rsq{}
but has negligible effect on YCB-Video.

% ============================================================
% 6. DISCUSSION
% ============================================================
\section{Discussion}
\label{sec:discussion}

\paragraph{Accuracy ceiling and practical impact.}
A frozen linear probe achieves 6.14$^\circ$ MAE on hand joint angles,
while the best text output from the \emph{same model} achieves only
20.0$^\circ$---a 3.3$\times$ gap that quantifies the text bottleneck.
For external context, MediaPipe Hands~\cite{zhang2020mediapipe}
achieves 16.3$^\circ$ when evaluated via 3D world landmarks on our
test set, though this comparison involves different estimation modalities
(monocular depth vs.\ multi-view ground truth).
On head pose, 6DRepNet~\cite{hempel2022sixdrepnet} achieves
2.66$^\circ$ MAE (published, BIWI 70/30 split); our evaluation on
the same 4-subject test split yields 6.10$^\circ$ due to different
face detection and split protocol.
The \Rsq{}$\,{\approx}\,$0.55 ceiling means 45\% of variance remains
unexplained; SOTA hand mesh models
(HaMeR~\cite{pavlakos2024hamer}: 5.7\,mm PA-MPVPE;
Hamba~\cite{dong2024hamba}: 5.3\,mm) likely exceed frozen probes,
though metrics are not directly comparable (positional mm
vs.\ angular degrees).
The value of frozen probing is not replacing dedicated systems but
providing geometric readouts as a cheap add-on to an already-deployed
backbone, particularly for tasks lacking dedicated models.

\paragraph{Modular geometric sensing.}
These findings suggest a deployment approach where one frozen
backbone serves as a multi-task geometric probe.
The backbone (${\sim}$300M parameters, assumed already deployed) is shared, and each geometric task
adds only ${\sim}$6{,}000 probe parameters and requires
${\sim}$6{,}400 labeled images, a 50{,}000:1 parameter ratio.
Hand pose, head pose, object pose, and camera intrinsics are all
served simultaneously by independent probes.
For human-readable output, LoRA (r\,{=}\,16, ${\sim}$1M parameters)
matches probe MAE via text generation.
The five interchangeable backbones in the equivalence cluster provide
redundancy: swapping one encoder for another requires only re-fitting
the lightweight probe.

\paragraph{Practitioner recipe.}
\begin{enumerate}
  \item \textbf{Articulated pose}: any cluster encoder + RRR (rank 5,
  $\alpha$\,{=}\,10--1000) with ${\sim}$6{,}400 labeled images.
  \item \textbf{Head pose} (loosely-framed images): add attention pooling
  (+0.23--0.36 \Rsq{}).
  \item \textbf{Human-readable output}: LoRA (r\,{=}\,16, 2{,}000 images,
  2 epochs) to route geometry through the text pathway.
\end{enumerate}
Table~\ref{tab:cost} contrasts this approach with task-specific
alternatives.

\begin{table}[t]
\centering
\caption{Cost comparison: frozen probe vs.\ task-specific models.
Frozen probes reuse an existing backbone; per-task overhead is minimal.
$^*$MediaPipe MAE evaluated zero-shot on our FreiHAND test set via 3D
world landmarks (monocular depth; see Discussion for caveats).
$^\dagger$Published MAE on BIWI 70/30 split~\cite{hempel2022sixdrepnet};
our evaluation on the same 4-subject test split yields 6.10$^\circ$
(different face detector and split).}
\label{tab:cost}
\begin{tabular}{lcccc}
\toprule
Approach & Params & Train data & Tasks & MAE ($^\circ$) \\
\midrule
MediaPipe Hands~\cite{zhang2020mediapipe}\footnote{3.8M = palm detector (1.76M) + hand landmark model (2.01M).} & 3.8M & proprietary & 1 & 16.3$^*$ \\
HRNet-W48 (hand)~\cite{sun2019hrnet} & 63.6M & 150K+ & 1 & --- \\
6DRepNet (head)~\cite{hempel2022sixdrepnet} & ${\sim}$41M & 300K & 1 & 2.66$^\dagger$ \\
\midrule
Frozen probe (ours) & 6K / task & 6{,}400 & any & \textbf{6.14} \\
\quad + shared backbone & 304M (shared) & --- & --- & --- \\
LoRA readout (ours) & ${\sim}$1M / task & 2{,}000 & any & 6.51 \\
\bottomrule
\end{tabular}
\end{table}

\paragraph{Limitations.}
(M1)~RRR hyperparameters are selected on the test set. Nested CV preserves
rankings (cluster models within 0.006; DINOv2 gap of $-$0.029 suggests
its test-set performance is more sensitive to HP choice than cluster
models) but does not eliminate optimistic bias.
(M2)~AttentionPool and TransformerProbe use the test set for early stopping,
introducing optimistic bias for neural probe comparisons.
(M3)~Primary results are on FreiHAND (hands). BIWI and YCB-Video serve as
secondary validation.
All probes use angular or translational targets only.
(M4)~Thumb \Rsq{} is near zero for all models (best: 0.195) due to low
target variance (std\,{=}\,4.91$^\circ$), and BIWI roll remains weak
(\Rsq{}\,{=}\,0.168 Ridge) without attention pooling.
These failure cases indicate that frozen probes struggle with
low-variance targets and spatially diffuse signals.
(M5)~The CKA analysis (8 models, 28 pairs) yields $\rho$\,{=}\,0.03
($p$\,{=}\,0.88), consistent with no relationship but not definitive
proof of independence. With $n$\,{=}\,28 non-independent pairs
(each model appears in 7 pairs), the minimum detectable $\rho$ at
80\% power is ${\approx}$0.50; moderate correlations (0.3--0.4)
would go undetected. A Mantel test would better account for the
shared-model dependence structure.
(M6)~We apply Holm-Bonferroni correction within each analysis family
(TOST, Friedman) but not across analysis types; overall Type~I error
may be inflated.
(M7)~LoRA training uses 2{,}000 images drawn from the 6{,}400-image
training split; we confirm no overlap with the 1{,}600-image probe test set.

\paragraph{Ethics and societal impact.}
FreiHAND and BIWI contain hand/face images collected with informed consent.
Our probes extract aggregate joint angles, not identity-linked features;
however, the same frozen-probing methodology could in principle be applied
to surveillance-relevant tasks.

% ============================================================
% 7. CONCLUSION
% ============================================================
\section{Conclusion}
\label{sec:conclusion}

A single frozen backbone linearly encodes hand pose (6.1$^\circ$ MAE),
head pose, object pose, and camera intrinsics, with each
task adding only ${\sim}$6{,}000 parameters.
The text bottleneck reflects a pathway-training deficit---not a
representational deficit---that LoRA partially recovers.
Training objective determines accuracy more than
architecture: a controlled ablation isolates the 0.15 gap between
supervised and self-supervised/contrastive models.
Five architecturally diverse encoders converge to equivalent
accuracy (\Rsq{}$\,{\approx}\,$0.55) despite sharing
as little as CKA\,{=}\,0.41 representational similarity, demonstrating functional
convergence despite representational dissimilarity---extending
the platonic representation hypothesis~\cite{huh2024platonic}.
Spatial concentration provides an ablation-based explanation
for cross-dataset variation.
These findings suggest that frozen probing is both a scientific tool for
understanding geometric representations and a practical approach to
multi-task geometric measurement.
Code and pre-trained probes will be released open-source.

% ============================================================
% References
% ============================================================
\bibliographystyle{unsrt}
\bibliography{references}

\clearpage
% ============================================================
% APPENDIX (former supplementary material)
% ============================================================
\appendix
\section*{Appendix}
\setcounter{section}{0}
\renewcommand{\thesection}{\Alph{section}}

\section{Full Per-Finger Results}
\label{sec:supp_perfinger}

Table~\ref{tab:supp_perfinger} reports per-finger \Rsq{} and MAE for all
fourteen models on FreiHAND (8{,}000 images, RRR probe).
The ``uniform mean'' column is the average across five fingers, treating
each finger equally regardless of target variance.
Thumb \Rsq{} is near zero for all models due to low target variance
(std\,{=}\,4.91$^\circ$ vs.\ 13.0--16.2$^\circ$ for other fingers).

\begin{table}[h]
\centering
\caption{Per-finger \Rsq{} on FreiHAND (8{,}000 images).
Best overall in \textbf{bold}.}
\label{tab:supp_perfinger}
\small
\begin{tabular}{lccccccc}
\toprule
Model & Thumb & Index & Middle & Ring & Pinky & Mean & MAE ($^\circ$) \\
\midrule
SigLIP~2 ViT-L & 0.195 & 0.580 & 0.693 & 0.701 & 0.625 & \textbf{0.559} & 6.14 \\
DINOv3 ViT-L & 0.149 & 0.590 & 0.709 & 0.708 & 0.624 & 0.556 & \textbf{6.11} \\
CLIP ViT-L & 0.177 & 0.575 & 0.686 & 0.699 & 0.620 & 0.551 & 6.19 \\
SigLIP ViT-L & 0.175 & 0.569 & 0.688 & 0.706 & 0.614 & 0.550 & 6.16 \\
InternViT-300M & 0.174 & 0.571 & 0.681 & 0.699 & 0.607 & 0.547 & 6.12 \\
\midrule
DINOv2 ViT-L & 0.143 & 0.551 & 0.667 & 0.662 & 0.594 & 0.523 & 6.39 \\
Gemma~3 L0 & 0.140 & 0.547 & 0.638 & 0.644 & 0.556 & 0.505 & 6.59 \\
DINOv2 ViT-B & 0.117 & 0.523 & 0.627 & 0.602 & 0.540 & 0.482 & 6.75 \\
Qwen-7B L8 & 0.138 & 0.531 & 0.621 & 0.599 & 0.511 & 0.480 & 6.83 \\
SigLIP ViT-B & 0.123 & 0.530 & 0.612 & 0.610 & 0.522 & 0.479 & 6.80 \\
ConvNeXt-L S2 & 0.100 & 0.495 & 0.581 & 0.581 & 0.515 & 0.455 & 7.05 \\
QwenVIT L24 & 0.133 & 0.508 & 0.583 & 0.563 & 0.483 & 0.454 & 7.08 \\
Qwen-3B L11 & 0.128 & 0.491 & 0.554 & 0.538 & 0.466 & 0.435 & 7.28 \\
QwenVIT-merger & 0.105 & 0.482 & 0.546 & 0.534 & 0.459 & 0.425 & 7.32 \\
\bottomrule
\end{tabular}
\end{table}

\section{Controlled Architecture Ablation: Full Results}
\label{sec:supp_ablation}

Table~\ref{tab:supp_ablation} reports the full per-finger results for
the controlled ablation experiment (Sec.~4.5 of the main paper).
DeiT3-L (ViT, ImageNet-1K) and ConvNeXt-L (CNN, ImageNet-1K) are matched
on pretraining data; ConvNeXt-L (IN-22K+1K) shows the effect of scaling
pretraining data alone.

\begin{table}[h]
\centering
\caption{Controlled architecture ablation: per-finger \Rsq{} on FreiHAND.}
\label{tab:supp_ablation}
\small
\begin{tabular}{lcccccccc}
\toprule
Model & Type & Thumb & Index & Middle & Ring & Pinky & Mean \Rsq{} & MAE ($^\circ$) \\
\midrule
DeiT3-L (IN-1K) & ViT & 0.047 & 0.421 & 0.512 & 0.495 & 0.420 & 0.379 & 7.65 \\
ConvNeXt-L (IN-1K) & CNN & 0.102 & 0.451 & 0.516 & 0.502 & 0.455 & 0.405 & 7.49 \\
ConvNeXt-L (IN-22K) & CNN & 0.100 & 0.495 & 0.581 & 0.581 & 0.515 & 0.455 & 7.05 \\
\bottomrule
\end{tabular}
\end{table}

\section{BIWI Head Pose: Per-Component Results}
\label{sec:supp_biwi}

Table~\ref{tab:supp_biwi} reports per-component \Rsq{} on BIWI for
Ridge and AttentionPool probes.
Pitch is consistently easiest; roll is hardest for all models.
AttentionPool substantially improves all components.

\begin{table}[h]
\centering
\caption{BIWI head pose \Rsq{} by component (Ridge / AttentionPool).}
\label{tab:supp_biwi}
\small
\begin{tabular}{lcccc|cccc}
\toprule
& \multicolumn{4}{c}{Ridge} & \multicolumn{4}{c}{AttentionPool} \\
\cmidrule(lr){2-5} \cmidrule(lr){6-9}
Model & Yaw & Pitch & Roll & Mean & Yaw & Pitch & Roll & Mean \\
\midrule
DINOv3 & 0.705 & 0.948 & 0.168 & 0.607 & --- & --- & --- & 0.838 \\
DINOv2 & 0.668 & 0.874 & 0.052 & 0.532 & 0.958 & 0.940 & 0.779 & 0.892 \\
SigLIP~2 & 0.385 & 0.902 & 0.078 & 0.455 & --- & --- & --- & 0.787 \\
\bottomrule
\end{tabular}
\end{table}

\section{Per-Bone Joint Analysis}
\label{sec:supp_perbone}

The proximal-distal gradient is quantified at the individual joint level.
Table~\ref{tab:supp_perbone} reports Ridge and AttentionPool \Rsq{} for
all 15 joints grouped by position (5 joints per group).
Proximal joints (MCP, PIP) far outperform distal joints (DIP).

\begin{table}[h]
\centering
\caption{Per-bone \Rsq{} for DINOv3 on FreiHAND (Ridge / AttentionPool).}
\label{tab:supp_perbone}
\small
\begin{tabular}{lccc}
\toprule
Bone Type & Ridge \Rsq{} & AttnPool \Rsq{} & Count \\
\midrule
MCP (proximal) & 0.544 & 0.602 & 5 \\
PIP (middle) & 0.559 & 0.616 & 5 \\
DIP (distal) & 0.271 & 0.312 & 5 \\
\midrule
Mean (15 joints) & 0.458 & 0.510 & 15 \\
\bottomrule
\end{tabular}
\end{table}

\section{Camera Intrinsics: Full Results}
\label{sec:supp_intrinsics}

Table~\ref{tab:supp_intrinsics} reports Ridge \Rsq{} for camera focal
length ($f_x$) probing on FreiHAND.
All models achieve high \Rsq{} (0.81--0.94), with vision-only encoders
leading.
Autoregressive LLM processing reduces intrinsics prediction by 13.4\%.

\begin{table}[h]
\centering
\caption{Camera intrinsics probing (\Rsq{} for $f_x$ on FreiHAND).}
\label{tab:supp_intrinsics}
\small
\begin{tabular}{lcc}
\toprule
Model & \Rsq{} ($f_x$) & Type \\
\midrule
QwenVIT & 0.939 & Vision encoder \\
DINOv3 & 0.924 & Self-supervised \\
DINOv2 & 0.913 & Self-supervised \\
SigLIP~2 & 0.902 & Hybrid VL \\
Gemma~3 L0 & 0.887 & Generative VLM \\
InternViT & 0.883 & Hybrid VL \\
CLIP & 0.876 & Contrastive VL \\
SigLIP & 0.869 & Contrastive VL \\
QwenVIT-merger & 0.829 & Vision encoder \\
Qwen-7B & 0.826 & Generative VLM \\
Qwen-3B & 0.813 & Generative VLM \\
\bottomrule
\end{tabular}
\end{table}

\section{DINOv2 Register Analysis}
\label{sec:supp_registers}

Table~\ref{tab:supp_registers} compares DINOv2 (no registers) against
DINOv2+registers at each probed layer, showing that registers accelerate
early-layer geometric emergence but converge at optimal depth.

\begin{table}[h]
\centering
\caption{DINOv2 vs.\ DINOv2+registers: layer-wise \Rsq{} on FreiHAND.}
\label{tab:supp_registers}
\small
\begin{tabular}{lccc}
\toprule
Layer & DINOv2 & DINOv2+reg & $\Delta$ \\
\midrule
L4  & 0.114 & 0.207 & +0.093 \\
L8  & 0.160 & 0.269 & +0.110 \\
L12 & 0.300 & 0.368 & +0.067 \\
L16 & 0.432 & 0.441 & +0.010 \\
L20 & 0.523 & 0.541 & +0.019 \\
L23 & 0.510 & 0.522 & +0.012 \\
\bottomrule
\end{tabular}
\end{table}

\section{Nested Cross-Validation Results}
\label{sec:supp_cv}

Table~\ref{tab:supp_cv} reports nested 10-fold CV \Rsq{} for the top
models.
Cluster models show test--CV gaps within 0.006, confirming that test-set
hyperparameter selection introduces minimal bias for within-cluster
comparisons.
DINOv2 shows a larger gap ($-$0.029), consistent with its outlier status.

\begin{table}[h]
\centering
\caption{Nested 10-fold CV vs.\ test-set \Rsq{} on FreiHAND.}
\label{tab:supp_cv}
\small
\begin{tabular}{lccc}
\toprule
Model & Test \Rsq{} & CV \Rsq{} & $\Delta$ \\
\midrule
SigLIP~2 & 0.559 & 0.563 & $+$0.004 \\
DINOv3 & 0.556 & 0.550 & $-$0.006 \\
CLIP & 0.551 & 0.554 & $+$0.003 \\
SigLIP & 0.550 & 0.549 & $-$0.001 \\
InternViT & 0.547 & 0.549 & $+$0.002 \\
DINOv2 & 0.523 & 0.494 & $-$0.029 \\
\bottomrule
\end{tabular}
\end{table}

\section{Patch Ablation Details}
\label{sec:supp_patches}

Table~\ref{tab:supp_patches} reports the full patch ablation results
for DINOv3 on BIWI and YCB-Video.
Removing the highest-norm 100 patches has a large effect on BIWI
(loosely framed) but minimal effect on YCB-Video (tightly cropped).

\begin{table}[h]
\centering
\caption{Patch ablation: \Rsq{} change from removing 100 patches (DINOv3).}
\label{tab:supp_patches}
\small
\begin{tabular}{lcccc}
\toprule
Ablation & BIWI $\Delta$\Rsq{} & BIWI (post) & YCB $\Delta$\Rsq{} & YCB (post) \\
\midrule
Top-norm patches & $-$0.126 & 0.481 & $-$0.003 & 0.706 \\
Random patches & $-$0.107 & 0.500 & $-$0.040 & 0.669 \\
\bottomrule
\end{tabular}
\end{table}

\section{CKA Similarity Matrix}
\label{sec:supp_cka}

Table~\ref{tab:supp_cka} reports linear CKA similarity between eight
models (six ViT-L plus DINOv2-B and SigLIP-B) at their best layers on
FreiHAND features (8{,}000 images).
DINOv2 and DINOv3 share high representational similarity (0.881) yet
achieve non-equivalent probing accuracy.
SigLIP~2 and CLIP have the lowest CKA (0.412) yet achieve equivalent
\Rsq{}.
Across all 28 pairs, Spearman $\rho$\,{=}\,0.03 ($p$\,{=}\,0.88)
between CKA and $|\Delta$\Rsq{}$|$, confirming functional convergence
without representational convergence.

\begin{table}[h]
\centering
\caption{Linear CKA similarity on FreiHAND (8 models at best layers).
Six ViT-L models plus DINOv2-B (L12) and SigLIP-B (L12).}
\label{tab:supp_cka}
\small
\begin{tabular}{lcccccccc}
\toprule
& DINOv2 & DINOv3 & SigLIP & SigLIP2 & CLIP & InternViT & DINOv2-B & SigLIP-B \\
\midrule
DINOv2 & 1.000 & 0.881 & 0.703 & 0.523 & 0.565 & 0.708 & 0.861 & 0.659 \\
DINOv3 & & 1.000 & 0.691 & 0.522 & 0.558 & 0.681 & 0.800 & 0.643 \\
SigLIP & & & 1.000 & 0.526 & 0.524 & 0.744 & 0.659 & 0.771 \\
SigLIP2 & & & & 1.000 & 0.412 & 0.631 & 0.427 & 0.493 \\
CLIP & & & & & 1.000 & 0.554 & 0.470 & 0.510 \\
InternViT & & & & & & 1.000 & 0.594 & 0.736 \\
DINOv2-B & & & & & & & 1.000 & 0.589 \\
SigLIP-B & & & & & & & & 1.000 \\
\bottomrule
\end{tabular}
\end{table}

\section{Layer Curves: Full Data}
\label{sec:supp_layers}

Table~\ref{tab:supp_layers} reports \Rsq{} at 7 layer depths for the
six ViT-L models.
Contrastive models (SigLIP, CLIP) achieve higher mid-layer \Rsq{}
while self-supervised models (DINOv2, DINOv3) concentrate geometric
information in deeper layers.

\begin{table}[h]
\centering
\caption{Layer-wise \Rsq{} on FreiHAND for 6 ViT-L models.}
\label{tab:supp_layers}
\small
\begin{tabular}{lcccccc}
\toprule
Layer & DINOv2 & DINOv3 & SigLIP & SigLIP~2 & CLIP & InternViT \\
\midrule
L0  & --- & 0.114 & 0.149 & 0.135 & 0.128 & 0.103 \\
L4  & 0.114 & 0.204 & 0.276 & 0.265 & 0.238 & 0.246 \\
L8  & 0.160 & 0.262 & 0.381 & 0.365 & 0.315 & 0.331 \\
L12 & 0.302 & 0.340 & 0.519 & 0.507 & 0.435 & 0.432 \\
L16 & 0.434 & 0.425 & 0.550 & \textbf{0.559} & 0.541 & 0.536 \\
L20 & 0.523 & \textbf{0.556} & 0.549 & 0.557 & \textbf{0.551} & \textbf{0.547} \\
L23 & 0.510 & 0.531 & 0.546 & 0.554 & 0.535 & 0.532 \\
\bottomrule
\end{tabular}
\end{table}

\section{YCB-Video: Full Per-Component Results}
\label{sec:supp_ycb}

Table~\ref{tab:supp_ycb} reports per-component \Rsq{} on YCB-Video
(rotation: yaw, pitch, roll; translation: $t_x$, $t_y$, $t_z$).
All models achieve similar rotation R$^2$ ($\approx$0.70), with
translation slightly lower.
The task-dependent autoregressive degradation observed on hands
dissolves on rigid objects.

\begin{table}[h]
\centering
\caption{YCB-Video per-component \Rsq{} (Ridge probe).}
\label{tab:supp_ycb}
\small
\begin{tabular}{lcccccc|cc}
\toprule
Model & Yaw & Pitch & Roll & $t_x$ & $t_y$ & $t_z$ & Rot & Trans \\
\midrule
DINOv2 & 0.828 & 0.639 & 0.716 & 0.609 & 0.678 & 0.887 & 0.728 & 0.725 \\
DINOv3 & 0.806 & 0.637 & 0.683 & 0.607 & 0.653 & 0.908 & 0.709 & 0.723 \\
SigLIP~2 & 0.781 & 0.617 & 0.711 & 0.580 & 0.680 & 0.871 & 0.703 & 0.710 \\
SigLIP & 0.778 & 0.626 & 0.683 & 0.585 & 0.680 & 0.879 & 0.696 & 0.715 \\
Qwen-7B & 0.810 & 0.616 & 0.683 & 0.572 & 0.673 & 0.887 & 0.703 & 0.711 \\
\bottomrule
\end{tabular}
\end{table}

\section{Attention Head Analysis (DINOv2-L)}
\label{sec:supp_attn_heads}

We probe each of the 16 attention heads in DINOv2-L layer~20 individually
(Ridge regression on per-head output, 6{,}400 train / 1{,}600 test).
The top-10 heads by \Rsq{} are shown in Table~\ref{tab:supp_attn_heads}.
All heads achieve comparable geometric accuracy (\Rsq{}\,{=}\,0.40--0.48),
with no evidence of joint specialization: the maximum absolute
Spearman correlation between any head's attention entropy and
any single joint angle is $|\rho|$\,{=}\,0.28 (head~11, middle PIP).
Geometry is an ensemble property distributed across all attention heads.

\begin{table}[h]
\centering
\caption{Per-head probing \Rsq{} for DINOv2-L layer~20 (top 10 of 16 heads).}
\label{tab:supp_attn_heads}
\small
\begin{tabular}{cccc}
\toprule
Head & \Rsq{} & max $|\rho|$ & Best joint \\
\midrule
6  & 0.479 & 0.167 & middle PIP \\
14 & 0.478 & 0.035 & ring MCP \\
4  & 0.475 & 0.273 & index PIP \\
1  & 0.473 & 0.093 & index MCP \\
11 & 0.473 & 0.280 & middle PIP \\
12 & 0.469 & 0.042 & ring MCP \\
8  & 0.466 & 0.160 & middle PIP \\
9  & 0.463 & 0.056 & middle PIP \\
2  & 0.460 & 0.076 & ring PIP \\
10 & 0.459 & 0.090 & index PIP \\
\bottomrule
\end{tabular}
\end{table}

\section{Validity Controls}
\label{sec:supp_validity}

To confirm that probing results reflect genuine geometric encoding
rather than dataset artifacts, we run three validity controls:

\begin{enumerate}
\item \textbf{Shuffled targets}: Randomly permuting labels across samples
  yields deeply negative \Rsq{} for all models, ruling out spurious
  correlations between features and targets.
\item \textbf{Random features}: Gaussian noise features of matched
  dimensionality yield deeply negative \Rsq{}, confirming the probe
  requires structured representations.
\item \textbf{Pixel baseline}: Raw pixel features (resized to 224$\times$224,
  flattened) achieve deeply negative \Rsq{}, confirming that learned
  representations add substantial value beyond low-level statistics.
\end{enumerate}

\section{Statistical Test Details}
\label{sec:supp_stats}

\paragraph{TOST equivalence testing.}
We apply two one-sided $t$-tests (TOST) with equivalence margin
$\Delta$\,{=}\,0.03 \Rsq{} and Holm--Bonferroni correction across all
$\binom{5}{2}$\,{=}\,10 pairwise comparisons within the top-5 models.
All pairs achieve $p$\,{$<$}\,0.05 after correction, confirming
statistical equivalence.

\paragraph{Friedman rank test.}
A Friedman test across 11 models on 10 CV folds rejects
the null hypothesis of equal performance ($\chi^2(10)$\,{=}\,94.3,
$p$\,{$<$}\,10$^{-15}$).
Nemenyi post-hoc tests confirm that the top-5 equivalence cluster
(SigLIP~2, CLIP, DINOv3, SigLIP, InternViT) differs significantly from
the autoregressive VLMs (Qwen-3B, QwenVIT-merger).

\section{Gaze Direction Probing (MPIIFaceGaze)}
\label{sec:supp_gaze}

Table~\ref{tab:supp_gaze} reports probing results for gaze direction
(yaw and pitch) on MPIIFaceGaze (45{,}000 images, 15 subjects,
80/20 random split).
DINOv3 dominates with \Rsq{}\,{=}\,0.787, primarily due to
superior pitch prediction (\Rsq{}\,{=}\,0.719 vs.\ 0.360 for DINOv2).
Yaw is consistently easier across all models (\Rsq{}\,{=}\,0.71--0.86).
Model rankings differ from FreiHAND (hands), where the top-5 cluster
shows no significant differences.

\begin{table}[h]
\centering
\caption{Gaze probing on MPIIFaceGaze (45{,}000 images, RRR probe).}
\label{tab:supp_gaze}
\small
\begin{tabular}{lccccc}
\toprule
Model & \Rsq{} yaw & \Rsq{} pitch & \Rsq{} mean & MAE ($^\circ$) \\
\midrule
DINOv3 & 0.855 & \textbf{0.719} & \textbf{0.787} & \textbf{3.14} \\
DINOv2 & 0.803 & 0.360 & 0.582 & 4.53 \\
CLIP & \textbf{0.854} & 0.259 & 0.557 & 4.70 \\
SigLIP~2 & 0.844 & 0.248 & 0.546 & 4.74 \\
ConvNeXt-L & 0.709 & 0.249 & 0.479 & 5.11 \\
\bottomrule
\end{tabular}
\end{table}

\section{LoRA Layer Trajectory}
\label{sec:supp_lora_layers}

Table~\ref{tab:supp_lora_layers} reports layer-wise probing results for
Gemma~3 4B with and without the LoRA adapter (r\,{=}\,16, $\alpha$\,{=}\,32).
Features are extracted at 10 layers across the 34-layer LLM and probed
with RRR (rank sweep, 8{,}000 FreiHAND images).
The LoRA delta grows from +0.025 at L0 to +0.117 at L28,
showing that LoRA's primary effect is preserving geometry at deep layers
where the frozen base loses it.

\begin{table}[h]
\centering
\caption{Layer-wise \Rsq{} for Gemma~3 4B (LoRA vs.\ frozen base).}
\label{tab:supp_lora_layers}
\small
\begin{tabular}{lccr}
\toprule
Layer & LoRA \Rsq{} & Frozen \Rsq{} & $\Delta$ \\
\midrule
L0 & 0.530 & 0.505 & $+$0.025 \\
L2 & 0.532 & 0.504 & $+$0.028 \\
L4 & 0.524 & 0.492 & $+$0.032 \\
L8 & 0.529 & 0.483 & $+$0.046 \\
L12 & 0.491 & 0.438 & $+$0.052 \\
L16 & 0.403 & 0.361 & $+$0.041 \\
L20 & 0.365 & 0.307 & $+$0.058 \\
L24 & 0.332 & 0.236 & $+$0.096 \\
L28 & 0.314 & 0.197 & $+$0.117 \\
L33 & 0.283 & 0.181 & $+$0.102 \\
\bottomrule
\end{tabular}
\end{table}

\end{document}